\def\year{2018}\relax
\documentclass[letterpaper]{article} %DO NOT CHANGE THIS
\usepackage{aaai18}  %Required
\usepackage{times}  %Required
\usepackage{helvet}  %Required
\usepackage{courier}  %Required
\usepackage{url}  %Required
\usepackage{graphicx}  %Required
\usepackage{hhline}
\usepackage{makecell}
\usepackage[switch]{lineno}
\begin{document}
\title{Video-based Person Re-Identification using Gated \\ Convolutional Recurrent Neural Networks}
\author{Yang Feng, Yu Wang, Jiebo Luo\\
University of Rochester \\
yfeng23@cs.rochester.edu
}
\maketitle
\begin{abstract}
Deep neural networks have been successfully applied to solving the video-based person re-identification problem with impressive results reported. The existing networks for person re-id are designed to extract discriminative features that preserve the identity information. Usually, whole video frames are fed into the neural networks and all the regions in a frame are equally treated. This may be a suboptimal choice because many regions, e.g., background regions in the video, are not related to the person. Furthermore, the person of interest may be occluded by another person or something else. These unrelated regions may hinder person re-identification. In this paper, we introduce a novel gating mechanism to deep neural networks. Our gating mechanism will learn which regions are helpful for person re-identification and let these regions pass the gate. The unrelated background regions or occluding regions are filtered out by the gate. In each frame, the color channels and optical flow channels provide quite different information. To better leverage such information, we generate one gate using the color channels and another gate using the optical flow channels. These two gates are combined to provide a more reliable gate with a novel fusion method. Experimental results on two major datasets demonstrate the performance improvements due to the proposed gating mechanism.
\end{abstract}

\section{Introduction}

%Many public places are equipment with surveillance cameras nowadays. Video streams are recorded twenty four hours per day and seven days per week by these cameras.When we are interested in someone, it is labor expensive to manually search through all the videos for him.
Person re-identification (re-id) aims at automatically matching the pedestrians across non-overlapping cameras. It is an important research topic as it can significantly reduce human labor when searching for someone in surveillance videos. Person re-id is also a very challenging problem because the variations in view angle, pose, and lighting can have a person look quite different under different camera views.

Person re-id has been extensively studied in the recent decade. Most existing schemes focus on image-based person re-id and many promising results have been achieved. Varior et al. \cite{varior2016gated} proposed a gating function to select the common local patterns between a pair of input images. In \cite{zhao2013unsupervised,zhao2017person}, the salience regions were estimated without supervision. The authors believed that salient regions provide valuable information for person re-id and saliency is also reliable across different camera views. Zhao et al. \cite{zhao2017spindle} detected several body regions and matched the corresponding regions of different samples. The motivation of this paper is related to these methods. We develop a gating mechanism to find the regions of the person of interest in videos. Our method is different from them since temporal information is used to find the person.

\begin{figure*}
\centering
\includegraphics[width=6in]{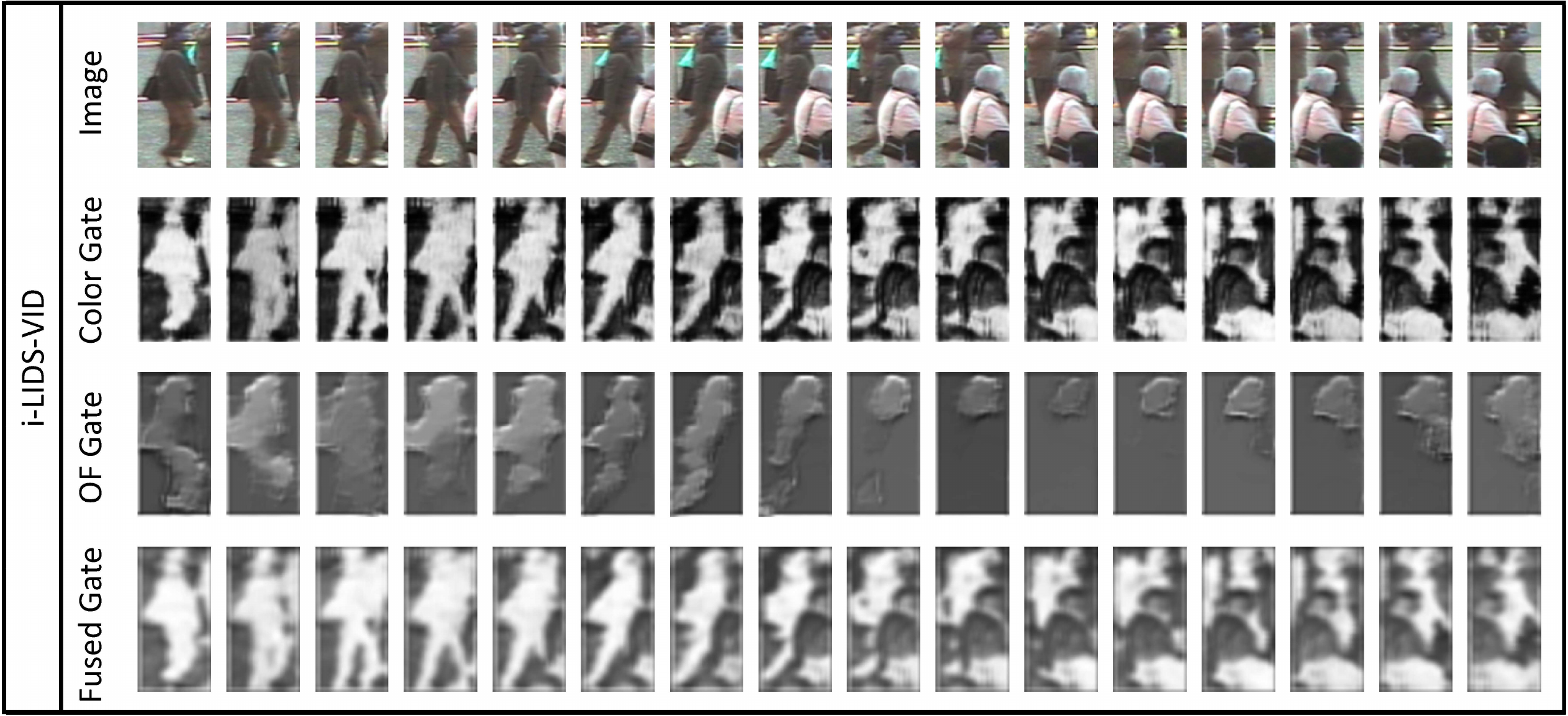}
\caption{The four rows from top to bottom show the original image, color gate, optical flow gate and fused gate, respectively.}
\label{gate}
\end{figure*}

Recently, several video-based person re-id methods have been proposed. In \cite{mclaughlin2016recurrent}, frames and optical flow were fed into a convolutional neural network (CNN) to extract frame features, which were further fed into a recurrent neural network (RNN). The outputs of the RNN at different time steps were combined as sequence features using temporal pooling. The whole network was trained in a Siamese fashion with an additional identity loss. Yan et al. \cite{yan2016person} extracted LBP and color features for each frame and aggregated them using LSTM \cite{hochreiter1997long}.
Instead of using the pre-extracted optical flow, Liu et al. \cite{liu2017video} generated optical flow using a motion network based on FlowNet \cite{fischer2015flownet}. After pre-training the motion network, they trained the whole model end-to-end. In \cite{zhou2017see,xu2017jointly}, temporal attention models were used to find the important frames for person re-id.

In \cite{mclaughlin2016recurrent,liu2017video}, the whole frames were used to extract features without differentiating the person of interest from unrelated regions. In a video sequence, the person of interest is usually located at the center of each frame. Inevitably, there are some background regions around the person in the video. The color information and optical flow in the background regions are distractive for extracting discriminative features for person re-id. In addition, the person of interest may be occluded by someone or something else. The occluding objects are also distractions to the person re-id task.

In the literature, visual attention mechanism has been used to find the important regions to examine further. In image captioning, an attention-based model was used to learn the corresponding region in the image to each word \cite{xu2015show}. In action recognition, an attention model was used to learn which parts in the frames are relevant to the task at hand and attach higher importance to them \cite{sharma2015action}.
%The video based person re-id problem has something in common with the action recognition problem. In both problems, there are people moving in the video. The action recognition problem focuses on what action the person is performing, while the person re-id problem focuses on who is moving around in the video.
We have considered applying the visual attention mechanism to person re-id but eventually gave it up because the resolution of the videos is too low.
With an attention model, the input frame is first fed into a pre-trained CNN, i.e. VGG16 \cite{simonyan2014very}. The feature vectors at the last convolutional layer are weighted and summed together to form a context vector.
%We cannot build a good baseline method for video-based person re-id problem using a pre-trained CNN yet.
The resolution of the videos for person re-id is usually low, i.e. $64 \times 128$. After downsampling in a pre-trained CNN, the width of the feature map becomes very small. It is difficult to distinguish the person and the background at that scale.

In this paper, we instead develop a gating mechanism to find the regions that are helpful for person re-id in the video frames. Compared with the attention mechanism, a gating mechanism is able to select regions at a fine-grained scale. Depending on which layer in the neural network we insert the gating mechanism to, the generated gate could be as large as the input image or several times smaller. It is possible to modulate the size of the gate so that it is suitable for the person re-id task.
%Gating mechanism has already used in image based person re-id \cite{varior2016gated}. In \cite{varior2016gated}, a gating function is used to selective the common local patterns between a pair of input images. We hope the the gate is able to select the regions of the person we are interested in.
To fulfill the goal of the gating mechanism, both the color channels and optical flow channels are used to generate the gate. The color channels and optical flow channels provide very different information, so we choose to generate one color gate and one optical flow gate and fuse them to obtain the final gate. Optical flow can roughly provide the information we need to generate the desired gate. During any short period, the person usually moves at a constant velocity, so the optical flows of the background regions are in a similar direction and magnitude. The optical flows from the person of interest are quite different from those of the background and occluding objects. The torso will be relatively stable and the limbs will move periodically. The occluding object usually moves in a different direction or at a different speed. The color channels also provide valuable information for deciding which region is useful. The background captured by one camera usually looks quite similar. The color gate can learn how to distinguish the background and the foreground. It is also possible that the color gate learns to catch some salient visual characteristics of a person, which are beneficial for person re-id.

Although the color gate and optical flow gate can catch some helpful regions, neither of them is perfect. The optical flow estimated from the surveillance videos is not accurate, thus the optical flow gate may also make some errors. If a person is wearing clothes similar to the background, the clothes may be filtered out by the color gate. To mitigate these problems, we design a fusion method to fuse the color gate and the optical gate so that the fused gate is more reliable. If one gate believes some regions are useful, the information in those regions will pass the gate. If neither of the gates thinks a region is useful, the information in that region will not pass the gate. Fig. \ref{gate} shows some examples of the generated gates. In the third row of Fig. \ref{gate}, we can see that the optical flow gate filters out the occluding person. Our fusion method is still not perfect because a region will pass the fused gate if either gate allows the region to pass by error, but it is beneficial to person re-id since it is able to keep the important regions while filtering out some distracting regions.

Designing a proper fusion method is nontrivial. It is desired that one gate should not hamper the learning of the other gate. For example, if one gate thinks a region is important, the other gate can still learn that the region is important. We also want to keep the values in the fused gate at the same scale, which means that the values of the selected regions should be similar and the values of the unselected regions should also be similar. Based on these two intuitions, we design a novel fusion function to achieve our goals.
%So we are able to select the regions helpful for person re-id using optical flow. However,  If we only use the optical flow to generate the gate, some useful information will lost. Without the lost information, the re-id performance may decrease. To solve this problem, we also generate a gate using the color channels and fuse these two gates by a function similar to OR operation. So 

Our contributions can be summarized as follows:
\begin{enumerate} %[1.]
	\item We introduce a novel gating mechanism to video-based person re-identification. The gating mechanism is able to select the regions that are useful for distinguishing the person of interest and filtering out the distracting regions.
	\item We propose a novel fusion method to integrate the color gate and optical flow gate so that the fusion of two gates will not hamper the learning of either gate.
	\item We conduct ablation experiments to examine the effect of each component of our method. The experimental results show that the performance of our gated network is better than non-gated networks.
\end{enumerate}

\section{Related Work}
\subsection{Image-based Person Re-ID}
Image-based person re-id has been extensively studied in recent years. For this problem, there is one image sample available for a person under per camera view. We want to match the images of the same person across different cameras. To achieve this goal, some image features are extracted first and the similarity of the images are calculated by a distance metric based on the image features. %In the early work by Gheissari et al. \cite{gheissari2006person} in 2006, a spatial-temporal segmentation method was used to detect the foreground clothing and group the pixels that belong to the same type of fabric into regions. Each region was represented by invariant signatures including normalized color histograms and salient edgel histograms. Gray and Tao \cite{gray2008viewpoint} designed a feature space consisting of 8 color channels and 21 texture filters intuitively. Adaboost algorithm was applied to the feature space to learn object class specific representation and a discriminative recognition model.
%In \cite{matsukawa2016hierarchical}, the mean information was added to a hierarchical covariance descriptor for better discriminate power in person re-id. Local regions were described by multiple Gaussian distributions of color and texture cues. A set of regions was described by another distribution.
%Some attributed-based mid-level features were also proposed for the person re-id problem because attributes are believed to be stable across cameras. Su et al. \cite{su2015multi} introduced a low-rank attribute embedding so that incorrect and incomplete attributes can be recovered to better describe people.
%A good distance metric is critical for the success of person re-id using hand-crafted features. Currently, KISSME \cite{koestinger2012large} is the most popular metric learning method for person re-id. A Mahalanobis distance is derived from the log-likelihood ratio test in KISSME.
The intuition of our gating mechanism is similar in spirit to some existing schemes  \cite{varior2016gated,zhao2017person,liu2016end}. Varior et al. \cite{varior2016gated} proposed a gating function to select the common local patterns between a pair of input images. This method is inefficient in large datasets because a probe image needs to be paired with each of the gallery images before being sent into the network. In \cite{zhao2017person}, Zhao et al. designed a method to learn the salience regions without supervision. Liu et al. \cite{liu2016end} introduced an attention mechanism to learn which parts of the image are relevant to discerning people. All these methods try to treat different regions in an image differently according to their relevance to distinguishing people. We are using the gating mechanism to achieve this goal and our gate is calculated using both spatial and temporal information.

\subsection{Video-based Person Re-ID}
Compared with image-based person re-id, the use of video for person re-id has several advantages. First, a video usually contains several frames, which means more samples are available. Second, the temporal information contained in videos is beneficial to person re-id. Some biometrics such as gait can be extracted using this temporal information. In \cite{farenzena2010person}, the gallery video and probe video were regarded as frame sets and no temporal information was used. Wang et al. \cite{wang2014person} designed a space-time feature representation of videos by combining HOG3D features and optical flow energy profile. In \cite{mclaughlin2016recurrent}, frames and optical flows were fed into a CNN to extract frame features, which were further fed into a RNN. The outputs of the RNN at different time steps were combined as sequence features using temporal pooling. The whole network was trained in a Siamese fashion with additional identity information.
%We select the network in \cite{mclaughlin2016recurrent} as our baseline and some designs in our network are similar.
The difference between our network and theirs is that there is a gating module in our network. And the color channels and optical flow channels are fed into our network separately.
%, instead of being concatenated and fed together.
% why ????????????? any advantages in doing differently???????
Instead of using pre-extracted optical flow, Liu et al. \cite{liu2017video} used a motion network based on FlowNet \cite{fischer2015flownet}. After pre-training the motion network, they trained the whole model end-to-end. Here we decided to use use pre-extracted optical flow because we are focusing on the gating mechanism in this paper. In \cite{yan2016person}, the manually defined feature vectors are aggregated by LSTM. The inputs of our model are not feature vectors but video frames and optical flow. 

\subsection{Convolutional Long Short Term Memory}
Long Short Term Memory (LSTM) \cite{hochreiter1997long} was proposed to learn the long term dependencies in sequential data. After changing the network structure by introducing cells and gates, the performance increases without changing the loss function. Shi et al. \cite{xingjian2015convolutional} developed convolutional LSTM (ConvLSTM) to model the spatiotemporal sequence in precipitation nowcasting. Wu et al. \cite{wu2016convolutional} applied ConvLSTM in an auto encoder fashion to learn video descriptors for person re-id. Our gating mechanism is motivated by the input gate in ConvLSTM. Convolutions are used in calculating the gates and hidden state in our network. The state and gate in ConvLSTM is a 3d cube, while the state is a 1d vector and the gate is a 2d map in our network.

\section{Methodology}
\begin{figure}
\centering
\includegraphics[width=3.5in]{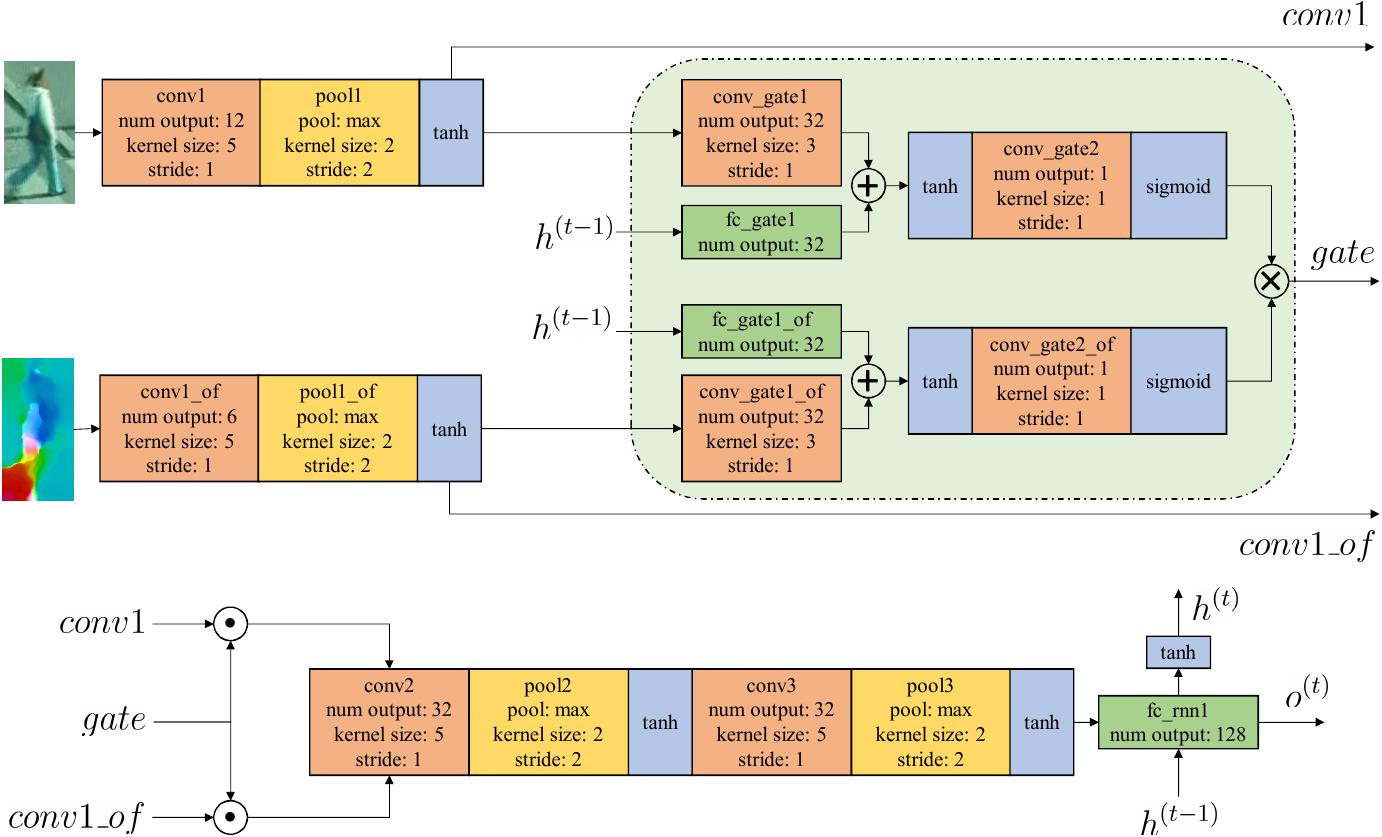}
\caption{An illustration of our model for feature extraction. The first line inside each box is its name. The boxes with names starting with ``conv'', ``pool'', and ``fc'' are convolutional, pooling and fully connected layers, respectively. The boxes named ``tanh'' or ``sigmoid'' are activation layers. $\bigoplus$ and $\bigodot$ means element-wise addition or multiplication. $\bigotimes$ is a fusion operation designed by ourselves. $h^{(t-1)}$ and $h^{(t)}$ are the RNN state at previous and current time step, respectively. The arrows show the data flowing direction. Some arrows between the convolutional layers, pooling layers and activation layers are omitted to save space.}
\label{framework}
\end{figure}

We introduce a novel gating mechanism into the recurrent convolutional network framework for video-based person re-id. The network structure for feature extraction is shown in Fig. \ref{framework}. To make our network structure clear, we draw all the convolutional, pooling, activation and fully-connected layers with their hyper-parameters written in them. The first row of the words in each layer is the name of that layer. We directly refer to a layer by its name in the following of this section. When we mention a convolutional layer by its name, the pooling layer and activation layer after that convolutional layer are included implicitly.

%Our network consists of a convolutional part and a recurrent part. Only ``conv1'' and ``conv1\_of'' belongs to the convolutional part. All the other layers belong to the recurrent part because the RNN state at previous time-step is used when calculating the gate.
Our network consists of two parts, namely the gate generation part and feature extraction part. The gate generation part and feature extraction part are shown in the top half and bottom half of Fig. \ref{framework}, respectively.
At each time step, our network takes two inputs: a frame and the corresponding optical flow. These two inputs first go through the first convolutional layers. The outputs of the first convolutional layers are fed into the gating module, which is marked by the light green region. In the gating module, a color gate and optical flow gate is calculated using the output of the first convolutional layer and the hidden state at the previous time step. After generating the two gates independently, they are fused into a new gate, which is sent to the feature extraction part together with the outputs of the first convolutional layer. In the feature extraction part, the outputs of the first convolution layers are filtered by the gate, concatenated, and further fed into another two convolutional layers and a RNN layer. The output of the RNN is the extracted feature for one frame. To obtain the feature for a video, we average all the features of the frames in the video.

In our network, all the pooling layers are $2\times 2$ max pooling layers. In all the convolutional layers, we pad the input so that the width and height of the output are the same with the input.

\subsection{Shared Convolutional Layers}
The layers ``conv1'' and ``conv1\_of'' are shared convolutional layers in both the gate generation part and feature extraction part. The processing steps for the color channels and the optical flow channels are very similar in the gate generation part. The differences are the number of input and output channels in the first convolutional layer. So we will only describe the processing steps for the color channels in this and the next subsections.

Suppose the width and height of the input frame is $W$ and $H$, respectively. The shape of the input X is $W\times H\times 3$. $conv1=tanh(pooling(conv(X)))$, where $conv1$ denotes the output of ``conv1'' layer. The shape of $conv1$ should be $\frac{W}{2}\times \frac{H}{2}\times 12$ if $W$ and $H$ are even numbers. After obtaining the output of ``conv1'' layer, we start calculating the gate.

\subsection{Gating Layers}
The gating module is contained in the light green region in Fig. \ref{framework}. Motivated by ConvLSTM \cite{xingjian2015convolutional} and attention model \cite{sharma2015action}, we use both the input at the current time step and the state at the previous time step to calculate the gate. After passing ``conv\_gate1'', the shape of the cube becomes $\frac{W}{2}\times \frac{H}{2}\times 32$. The state at the previous time step is a 128-dim vector, which is projected to a 32-dim vector by the fully-connected layer ``fc\_gate1''. $\bigoplus$ represents element-wise addition operation. When performing the element-wise addition, we need to broadcast the 32-dim vector to a  $\frac{W}{2}\times \frac{H}{2}\times 32$ cube so that the two inputs of $\bigoplus$ have the same shape. After the summation, tanh is applied to all the values in the cube. The output of that tanh layer is mapped into a one channel plane by ``conv\_gate2'' layer. In the end, the color gate is generated by a sigmoid activation function.

The optical flow gate is generated in a similar way.
%After generating both gates, we want to design a fusion function to fuse them so that that they can complement with each other.
Neither the color gate nor the optical flow gate is perfect for filtering out the harmful regions while keeping the helpful regions. Therefore, we design a method to fuse the color gate and the optical flow gate so that they can complement each other. Based on the observation that the accuracy will not drop when the fused gate let all regions pass, we intuitively want to keep more regions properly. We will keep a region when either the color gate or the optical flow gate believes the region is helpful. To fulfill our goal, our fusion function is defined as

\begin{equation}
\begin{array}{rl}
G_{fused} &= G_{color} \bigotimes G_{of} \\
&= G_{color} + G_{of} - [G_{color} \bigodot G_{of}]_{const},
\end{array}
\end{equation}

\noindent where $G_{color}$, $G_{of}$, and $G_{fused}$ denote the color gate, optical flow gate, and fused gate, respectively. $[\cdot]_{const}$ means treating the expression as a constant value. The variable in $[\cdot]_{const}$ is regarded as a constant and it will not produce any gradient during back-propagation. $\bigotimes$ and $\bigodot$ are our fusion function and the element-wise multiplication, respectively. The fusion function is designed with the following considerations. If either the color gate or the optical flow gate believes some regions are useful, the information at those regions is allowed to pass the gate. Therefore, we take the summation of two gates first. After summation, the scales of the values in the fused gate are inconsistent. The gate values of the regions that both gates agree to pass may be twice as large as the gate values of the regions where only one gate allows passing. To solve this problem, we subtract the element-wise multiplication of the two gates from the summation. A new problem arises after adding the subtraction term. If one gate believes a region is important, it will hamper the learning of the other gate. For example, the values in the color gate equal to one at some regions, where there will be no gradient generated for the optical flow gate for the same regions. To solve this new problem, we introduce the constant operation $[\cdot]_{const}$. $\frac{\partial G_{fused}}{\partial G_{color}}$ and $\frac{\partial G_{fused}}{\partial G_{of}}$ will always be one after adding the constant operation.

\subsection{Feature Extraction Layers}
After calculating the fused gate, we apply the gate to the outputs of both ``conv1'' layer and ``conv1\_of'' layer. $\bigodot$ in Fig. \ref{framework} is the element-wise multiplication operation. Because there is only one channel in the fused gate and there are several channels in the outputs of the convolutional layers, the fused gate is broadcast when doing the element-wise multiplication. After filtering by the fused gate, the outputs of ``conv1'' layer and ``conv1\_of'' layer are concatenated along the channel dimension.
%The design of the following layers are similar to the network structure in \cite{mclaughlin2016recurrent}.
The concatenated cube is fed into another two convolutional layers. The information in different frames is aggregated by a recurrent layer in the end.
The recurrent layer is given by
\begin{equation}
\begin{array}{c}

o^{(t)} = W\left(
\begin{array}{c}
conv3\\
h^{(t-1)}
\end{array}
\right)+b \\
%\end{equation}
%\begin{equation}
h^{(t)} = tanh(o^{(t)})
\end{array}
\end{equation}

\noindent where $conv3$ denotes the flattened output of the ``conv3'' layer. $h^{(t-1)}$ and $h^{(t)}$ are the RNN state at previous and current time step, respectively. $o^{(t)}$ is the output of the RNN at current time step. $W$ and $b$ are the parameters of the RNN. After we compute the feature of one frame, we use average pooling to incorporate the features in all the frames of a video.

\begin{figure}
\centering
\includegraphics[width=3.5in]{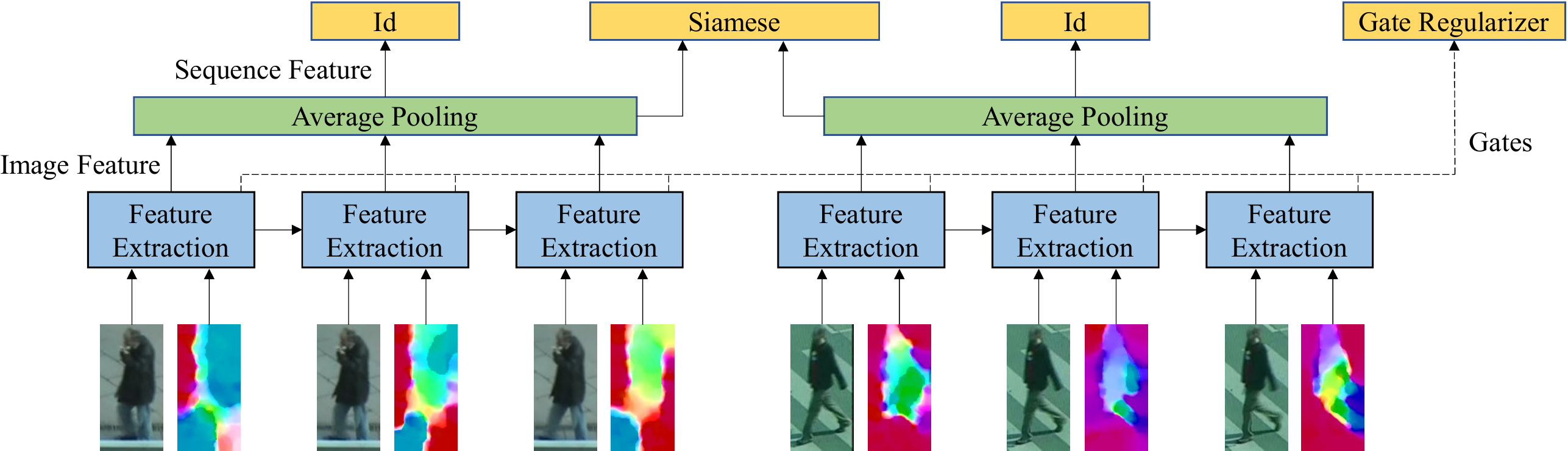}
\caption{An illustration of the training process. We train our network using the identification loss, Siamese loss and gate regularizer jointly.}
\label{fig:loss}
\end{figure}

\subsection{Network Training}
\label{loss}
The whole network training structure is illustrated in Fig. \ref{fig:loss}. Similar to \cite{mclaughlin2016recurrent}, we also use the identification loss and verification loss together as our loss function. The identification loss is used to penalize the wrong classification, while the verification loss encourages the distance of feature vectors of the same person to be small and the distance of the feature vectors of different persons to be large. Let $v_i$ and $v_j$ denote the feature vectors of the i-th and the j-th person, respectively. The probability that $v_i$ is a feature vector of the c-th person is
\begin{equation}
P(id=c|v_i)=\frac{\exp(W_cv_i)}{\sum_k\exp(W_kv_i)}
\end{equation}
where W is a weight matrix. The identification loss is defined as
\begin{equation}
L_{id}(v_i)=-log(P(id=i|v_i)).
\end{equation}

The verification loss is defined as
\begin{equation}
\label{siameseloss}
L_{ver}(v_i, v_j)=\left\{
\begin{array}{cc}
\frac{1}{2}\|v_i-v_j\|^2 & i=j \\
\frac{1}{2}\max(m-\|v_i-v_j\|, 0)^2 & i\neq j
\end{array}
\right.,
\end{equation}
where $m$ is the margin.

%\subsection{Gate regularizer}
If the network is trained only with the losses defined above, the values of the fused gate tend to be very small. It is because that the verification loss for the input pairs of the same person will dominate the training process after training for a while. The training process can be divided into two main steps. In the first step, the identity loss will dominate the training. We are using the negative log likelihood function as the loss function. It can produce a very large gradient when the probability of classifying a person correctly is small because the log function is very steep near zero. After training the network for a while, the accuracy of classification is almost 100\%. The identify loss reduces to a small number and the Siamese loss becomes the dominating one. At this moment, the distance between a randomly sampled negative pair of video features is always larger than the Siamese margin. So only the first part in Eq. \ref{siameseloss} is effective and the second part is always zero. The network is updated so that the distance between features becomes small. That is why the values of the fused gate become small after training the network for some epochs.

The decrease in the values in the fused gate is harmful to  training because the gradient of the sigmoid function is very small when the input is far from zero. To solve this problem, we add a gate regularizer:

\begin{equation}
\begin{array}{rl}
L_{gate}(G_{fused}) = &Max(0.5 - Mean(G_{fused}), 0) \\
&* (1 - Mean(G_{fused}))
\end{array}
\end{equation}

The intuition of this regularizer is that we want to set the mean value of the fused gate around 0.5. The regularizer will only penalize when the mean value of the fused gate is below 0.5. And the larger the distance between the mean value and 0.5 is, the more regularizer will penalize.

The overall loss of our network is an equally weighted sum of the identification loss, verification loss, and gate regularizer. It is defined as
\begin{equation}
\begin{array}{rl}
L = &L_{id}(v_i) + L_{id}(v_j) + L_{ver}(v_i, v_j) +\\
&\frac{1}{l_i}\sum_k^{l_i}L_{gate}(g_i^k) + \frac{1}{l_j}\sum_k^{l_j}L_{gate}(g_j^k)
\end{array}
\end{equation}
where $l_i$ and $l_j$ are the numbers of frames of the video used to generate $v_i$ and $v_j$, respectively. $g_i^k$ is the fused gate for the k-th frame of the video used to generate $v_i$. 

\begin{table*}
\small
\caption{The results of our network and other state-of-the-art methods.}
\label{table:res}
\centering
\begin{tabular}{|l|c|c|c|c|c|c|c|c|}
\hline
	Dataset & \multicolumn{4}{c|}{iLIDS-VID} & \multicolumn{4}{c|}{PRID-2011} \\ \hline
    Methods & Rank 1 & Rank 5 & Rank 10 & Rank 20 & Rank 1 & Rank 5 & Rank 10 & Rank 20 \\ \hline
    Baseline & 58.0 & 84.0 & 91.0 & 96.0 & 70.0 & 90.0 & 95.0 & 97.0 \\ \hline
    Baseline + EpicFlow & 58.8 & 83.5 & 91.4 & 96.2 & 70.9 & 90.7 & 95.0 & 98.6 \\ \hline
    \bf{Ours} & 62.8 & 87.1 & 92.5 & 98.2 & 75.0 & 93.0 & 96.4 & 99.0\\ \hhline{|=========|} %599 399
    AMOC \cite{liu2017video} & 68.7 & 94.3 & 98.3 & 99.3 & 83.7 & 98.3 & 99.4 & 100 \\ \hline
    ASTPN \cite{xu2017jointly} & 62.0 & 86.0 & 94.0 & 98.0 & 77.0 & 95.0 & 99.0 & 99.0 \\ \hline
    TAM + SRM \cite{zhou2017see} & 55.2 & 86.5 & - & 97.0 & 79.4 & 94.4 & - & 99.3 \\ \hline
    CNN + XQDA \cite{zheng2016mars} & 53.0 & 81.4 & - & 95.1 & 77.3 & 93.5 & - & 99.3 \\ \hline
    SI$^2$DL \cite{zhu2016video} & 48.7 & 81.1 & 89.2 & 97.3 & 76.7 & 95.6 & 96.7 & 98.9 \\ \hline
    RFA \cite{yan2016person} & 49.3 & 76.8 & 85.3 & 90.0 & 58.2 & 85.8 & 93.4 & 97.9 \\ \hline
    TDL \cite{You_2016_CVPR} & 56.7 & 80.0 & 87.6 & 93.6 & 56.3 & 87.6 & 95.6 & 98.3 \\ \hline
    DVR \cite{wang2016person} & 39.5 & 61.1 & 71.7 & 81.0 & 40.0 & 71.7 & 84.5 & 92.2 \\ \hline
    AvgTAPR \cite{gao2016temporally} & 55.0 & 87.5 & 93.8 & 97.2 & 68.6 & 94.6 & 97.4 & 98.9 \\ \hline
    STFV3D \cite{liu2015spatio} & 44.3 & 71.7 & 83.7 & 91.7 & 64.1 & 87.3 & 89.9 & 92.0 \\ \hline
\end{tabular}
\end{table*}

\section{Experiments}
In this section, we evaluate our person re-id method on two public video datasets: iLIDS-VID \cite{ilids} and PRID-2011 \cite{prid2011}. We first describe the datasets and experimental settings. Then the result of our method is reported and compared with other methods. In the end, we conduct several ablation studies to assess the contribution of each component of our method.

\subsection{Datasets}
\subsubsection{iLIDS-VID Dataset}
The iLIDS-VID person sequence dataset consists of image sequences captured by two non-overlapping camera views at an airport arrival hall. There are 300 different perons in iLIDS-VID dataset and each person has one image sequence under each camera view. So there are 600 image sequences in total. The lengths of the image sequences vary from 23 to 192. Because of clothing similarities among perons, occlusions, and viewpoint variations across camera views, person re-id on iLIDS-VID dataset is a very challenging task. 

\subsubsection{PRID-2011 Dataset}
In PRID-2011 dataset, there are 385 image sequences captured by one camera and 749 image sequences captured by another camera. The lengths of the image sequences in this dataset range from 5 to 675. The first 200 persons appear in both cameras. Following the protocol used in \cite{mclaughlin2016recurrent}, only the image sequences of the first 200 persons are used in our experiment. This dataset is less challenging than iLIDS-VID dataset because the background is relatively clean and there are few occlusions.

\subsection{Experimental Settings}
In our experiments, the image sequences of half of the persons are used for training and the image sequence of the other half are used for testing. For both datasets, we repeat our experiments for 10 times with different training/testing split and average the results to make the results more stable.
%Several hyper-parameters we use are the same with \cite{mclaughlin2016recurrent} so that we are more comparable. These parameters include t
The margin in the Siamese loss function, the feature embedding-space dimension and the learning rate are set to 2, 128 and 1e-3, respectively. We choose Adam \cite{kingma2014adam} as the optimization method. The parameters for Adam optimization method are left as defaults: $\beta_1=0.9$ and $\beta_2=0.999$.

%One parameter that is set different from \cite{mclaughlin2016recurrent} is the batch size. We set the batch size to 20 mainly because of speed issues. As the gate calculation is based on the RNN state at the previous time step, all the following layers must be computed one time step after another. Much less computation in layers ``conv2'' and ``conv3'' can be done in parallel after adding the gating mechanism. So the overall training process becomes slower. After increasing the batch size, the computation in ``conv2'' and ``conv3'' can be parallelized similar to the network without gating mechanism and the speed also becomes similar.
In each batch, there are 10 positive pairs and 10 negative pairs. The positive pairs are chosen as the image sequences of the same person under different camera views and the negative pairs are image sequences of different persons. When training the network, we do not use the whole image sequences because of computational reasons. We randomly choose sub-sequences with 16 consecutive frames as training samples. If the length of an image sequence is shorter than 16, we use that whole sequence. During testing, the first camera view is regarded as probe view and the second camera is regarded as gallery view. The maximum length of an image sequence allowed is set to 128 in testing time. Cumulative Match Characteristic (CMC) curve is used as the evaluation metric. We first calculate the feature vectors of all the samples in gallery set. For each probe sample, the Euclidean distance between its feature vector and all the feature vectors in gallery set are computed and sorted. The percentage of correct matches among the first m persons is regarded as CMC rank m.

All the images are converted to the CIE YUV color space. Optical flow is calculated between each pair of frames by the EpicFlow method \cite{epicflow}. There are three color channels and two optical flow channels. We compute the mean value and standard deviation of each channel on the whole training set. All the five channels are normalized to zero mean and unit variance. The same mean and standard deviation values are applied to the testing set. To augment the training data, we randomly crop a $120\times 56$ patch from the $128\times 64$ image. The patches are horizontally flipped with a probability of 0.5. When augmenting the testing data, we follow the protocols used in \cite{mclaughlin2016recurrent}. The distance between all the samples in gallery set and probe set are calculated eight times with different cropping offset. These distances are calculated again with all the patches flipped horizontally. We take the sum of all the distances calculated as the final distance for a pair of samples.

We implement our network by Tensorflow \cite{tensorflow2015-whitepaper}. The experiments are carried out using a Nvidia GTX 1070 GPU. It takes a day to finish the training on iLIDS-VID dataset for all training/testing splits.

\subsection{Results}
We report the results of our network in Table \ref{table:res} and first compare them with the results of the baseline network proposed in \cite{mclaughlin2016recurrent}.
%When implementing our system, we have many settings kept the same as \cite{mclaughlin2016recurrent} for fairness. The major setting differences include the batch size, optimization method, input image size, and the optical flow extraction method.
To demonstrate the effect of different implementations and replacing optical flow with EpicFlow, we carry out an experiment called ``Baseline + EpicFlow'', in which we implement the baseline method in Tenorflow and use EpicFlow instead of the optical flow generated by Lucas-Kanade algorithm \cite{lucas1981iterative}. The results of ``Baseline + EpicFlow'' is similar to the results reported in \cite{mclaughlin2016recurrent}.
%Different from the results reported in \cite{liu2017video}, the results of ``Baseline + EpicFlow'' we get are quite similar to the baseline method. This may be because there are some detailed differences between our implementation and Liu's.
By comparing the results of ``Baseline + EpicFlow'' and our method, we discover that our gating mechanism improves the baseline by a large margin. The improvement of the results demonstrate that filtering out the distracting regions is helpful to person re-id.

In the lower part of Table \ref{table:res}, the results of several state-of-the-art methods are listed as references. These methods include: AMOC \cite{liu2017video}, ASTPN \cite{xu2017jointly}, TAM + SRM \cite{zhou2017see}, CNN + XQDA \cite{zheng2016mars}, SI$^2$DL \cite{zhu2016video}, RFA \cite{yan2016person}, TDL \cite{You_2016_CVPR}, DVR \cite{wang2016person}, AvgTAPR \cite{gao2016temporally} and STFV3D \cite{liu2015spatio}.
Among all the methods, AMOC achieves the best results on both datasets because a motion network is designed to extract the motion information, which is better than optical flow. There is no gating or attention mechanism in AMOC and the whole frame is equally treated in it. We believe the results of AMOC could be further improved if our gating mechanism is incorporated into it.
On the iLIDS-VID dataset, the Rank 1 accuracy of our method is better than all the other methods except AMOC, which shows the advantage of suppressing background and occluding regions by our gating mechanism. We note that the Rank 1 accuracy of ``TAM+SRM'', ``ASTPN'', ``CNN + XQDA'' and ``SI$^2DL$'' are also higher than our method on the PRID-2011 dataset. This may be because the PRID-2011 dataset is relatively easy and occlusion rarely happens. The ability of filtering out occluding regions cannot take effect in this dataset. In addition, the CNN features used in \cite{zheng2016mars} are extracted by CaffeNet \cite{krizhevsky2012imagenet}, which are empirically more powerful than the CNN counterpart in our network.

\begin{table}
\small
\caption{The results of of the ablation experiments. ``[Color, OF]'' indicates the concatenation of the color and optical flow channels.}
\label{table:ablation}
\centering
\begin{tabular}{|c|c|c|c|c|}
\hline
	Dataset & \multicolumn{4}{c|}{iLIDS-VID} \\ \hline
    Methods & Rank 1 & Rank 5 & Rank 10 & Rank 20 \\ \hline
    Color Gate & 62.1 & 87.0 & 93.9 & 98.2 \\ \hline
    OF Gate & 59.2 & 85.5 & 91.7 & 96.5 \\ \hline
    [Color, OF] Gate & 62.7 & 85.8 & 92.5 & 97.2 \\ \hline
    Without RNN state & 60.2 & 85.9 & 93.1 & 97.7 \\ \hline
    Without Regularizer & 61.9 & 87.9 & 93.7 & 97.9 \\ \hline
    %Ours & 62.8 & 87.1 & 92.5 & 98.2 \\ \hline
\end{tabular}
\end{table}

\subsection{Ablation Experiments}
\label{ablation}
To analyze the contribution of each part of our gating mechanism, we conduct five ablation experiments: 1) generating the gate only using the color channels; 2) generating the gate only using the optical flow channels; 3) generating one gate using the concatenation of the color channels and optical flow channels; 4) removing the RNN state at the previous time state in gate calculation; 5) removing the gate regularizer. Because the iLIDS-VID dataset is more challenging, we only conduct the experiments on that dataset in this and next subsections. The results of these ablation experiments are listed in Table \ref{table:ablation}. From these results, we have the following observations:
\begin{itemize}
	\item Several gating methods produce better results than ``Baseline + EpicFlow'', which shows that the gating mechanism is helpful for person re-id.
	\item The gate generated only using optical flow is worse than others. This may be because the optical flow generated from surveillance videos is not very accurate. Some regions especially the regions of moving legs are often not visible in the optical flow. Although the optical flow gate is able to filter out some distracting regions, some valuable information for person re-id is also affected.
    \item The gate generated using the color channels or the concatenation of the color channels and optical flow channels is slightly worse than our fused gate.
	\item Calculating the gate without the RNN state at the previous time step causes performance drops. 
	\item The gate regularizer also improves the performance.
\end{itemize}

\subsection{Fusion Function Study}
We design several fusion functions and compare their performances to find the best one. %As showed in Section \ref{ablation}, the accuracy will drop if only using the color gate or only using the optical gate.
The  four designed fusion functions are shown in Table \ref{fuse_functions}. $Max(\cdot,\cdot)$ means taking the element-wise max and $\bigodot$ is the element-wise multiplication operation. All the fusion functions are designed following the intuition that we will let some data pass if either the color gate or the optical flow allows the data to pass. The results of different fusion function are displayed in Table \ref{table:fuse_res}. We observe that the fourth fusion function achieves the best results. The problem of the first fusion function is that the scale of the values in the fused gate is not consistent. Some of the gate values for the regions which are allowed to pass the gate may be twice as large  as the others. This inconsistent scaling of the data will make the layers behind the gate difficult to learn. Using the second and third fusion functions,  the learning of one gate may hamper the learning of the other gate. For example, if the values in the color gate are all the ones at a certain moment, the learning of the optical flow gate will stop at that moment. This is because the derivative of the fused gate with respect to the optical flow gate would be zero. The fourth fusion function overcomes the weakness of the other three, so it achieves the best results.
\begin{table}
\small
\caption{Four different fusion functions.}
\label{fuse_functions}
\centering
\begin{tabular}{|c|c|}
\hline
	No. & fusion Function \\ \hline
    1 & $G_{f1} = G_{color} + G_{of} $\\ \hline
    2 & $G_{f2} = Max(G_{color}, G_{of})$\\ \hline
    3 & $G_{f2} = G_{color} + G_{of} - G_{color} \bigodot G_{of}$ \\ \hline
    4 & $G_{f3} = G_{color} + G_{of} - [G_{color} \bigodot G_{of}]_{const}$ \\ \hline
\end{tabular}
\end{table}

\begin{table}
\small
\caption{The results of of using different fusion functions.}
\label{table:fuse_res}
\centering
\begin{tabular}{|c|c|c|c|c|}
\hline
	Dataset & \multicolumn{4}{c|}{iLIDS-VID} \\ \hline
    fusion Function & Rank 1 & Rank 5 & Rank 10 & Rank 20 \\ \hline
    $G_{f1}$ & 61.3 & 87.0 & 93.3 & 97.7 \\ \hline
    $G_{f2}$ & 59.3 & 85.6 & 92.3 & 96.7 \\ \hline
    $G_{f3}$ & 62.0 & 86.4 & 92.8 & 97.5 \\ \hline
    $G_{f4}$ & 62.8 & 87.1 & 92.5 & 98.2 \\ \hline
\end{tabular}
\end{table}

\section{Conclusions}
In this paper, we have introduced a novel gating mechanism into deep neural networks for video-based person re-id. The gating mechanism can learn which regions in a frame are useful for person re-id and which regions are distractions. By only letting the information from the helpful regions pass the gate, the gating mechanism improves the person re-id performance. To better leverage the information in the color channels and optical flow channels, a color gate and an optical flow gate are first generated independently. These two gates are fused to the final gate by one function, which lets some data pass if either color gate or the optical flow gate allows the data to pass. Experiments on two datasets have demonstrated the effectiveness of our gating mechanism.

%References and End of Paper
%These lines must be placed at the end of your paper
\bibliography{aaai.bib}
\bibliographystyle{aaai}
\end{document}